\documentclass{IEEEtran}
\usepackage{amsmath}
\usepackage{amssymb}
\usepackage{algorithmic}
\usepackage{array}
\usepackage{fixltx2e}
\usepackage{algorithm}
\usepackage{bm}
\usepackage{graphicx}
\usepackage{fancyvrb}

\usepackage[utf8]{inputenc}

\usepackage[noadjust]{cite}

\usepackage{color,soul}
\soulregister\cite7
\soulregister\ref7
\soulregister\pageref7

\begin{document}

\newcommand{\forecastexoCostk}{\hat{\x}_{k,\mathrm{ex}}^{\mathrm{c}}}

\newcommand{\timing}{\mathrm{time}}
\newcommand{\physical}{\mathrm{phys}}
\newcommand{\exo}{\mathrm{exo}}

\newcommand{\topLayer}{|\mathcal{L}|}

\newcommand{\Ta}{T^{\mathrm{air}}_{k}}
\newcommand{\Tm}{T^{\mathrm{env}}_{k}}
\newcommand{\Tout}{T^{\mathrm{out}}}
\newcommand{\Texo}{T^{\mathrm{\exo}}_{k}}

\newcommand{\xtiming}{x^{\timing}}
\newcommand{\xtimingnext}{x^{\timing'}}
\newcommand{\xtimingk}{x^{\timing}_k}
\newcommand{\xtimingt}{x^{\timing}_t}

\newcommand{\sens}{\mathrm{l}}

\newcommand{\x}{\bm{x}}
\newcommand{\augx}{\x^{\mathrm{aug}}}
\newcommand{\obsx}{\x^{\mathrm{obs}}}
\newcommand{\obsxNext}{\x^{\mathrm{obs}'}}
\newcommand{\augxNext}{\x^{\mathrm{aug}'}}
\newcommand{\obsk}{\bm{o}^{\physical}_k}
\newcommand{\obsl}{\bm{o}^{\physical}_l}
\newcommand{\obs}{\bm{o}^{\physical}}

\newcommand{\xhatexo}{\hat{\x}^{\exo}}
\newcommand{\xhatexol}{\hat{\x}^{\exo}_{l}}
\newcommand{\xexok}{\bm{x}^{\exo}_{k}}
\newcommand{\xexo}{\bm{x}^{\exo}}
\newcommand{\xexol}{\bm{x}^{\exo}_{l}}
\newcommand{\xhatexok}{\hat{\x}^{\exo}_{k}}
\newcommand{\xhatexokNext}{\hat{\x}^{\exo'}_{k}}
\newcommand{\xhatexolNext}{\hat{\x}^{\exo'}_{l}}

\renewcommand{\algorithmicrequire}{\textbf{Input:}}
\renewcommand{\algorithmicensure}{\textbf{Output:}}
\newcommand{\xblprime}{\bm{x}^{\mathrm{b}\prime}_{l}}

\newcommand{\uphys}{u^{\physical}}
\newcommand{\uphysl}{\uphys_{l}}
\newcommand{\uphk}{\uphys_{k}}
\newcommand{\uphysk}{\uphys_{k}}

\newcommand{\probwk}{p_{\mathcal{\omega}}(\cdot)}
\newcommand{\expected}{\mathbb{E}}
\newcommand{\Uphys}{U^{\physical}}

\newcommand{\histx}{\x^{\mathrm{hist}}}
\newcommand{\histxnext}{\x^{\mathrm{hist}'}}
\newcommand{\xphys}{\x^{\physical}}
\newcommand{\flow}{\dot{m}}
\newcommand{\xlstm}{\bm{x}}




\title{{Direct Load Control of Thermostatically Controlled Loads Based on Sparse Observations Using \\  Deep Reinforcement Learning}}

\author{Frederik~Ruelens, 
        Bert~J.~Claessens,
		Peter~Vrancx, 
		Fred~Spiessens,
		and Geert~Deconinck 

\thanks{F. Ruelens and G. Deconinck are with the Department of Electrical
	Engineering, KU Leuven/EnergyVille, 3000 Leuven, Belgium ({frederik.ruelens, geert.deconinck@kuleuven.be}).}
\thanks{\text{B.~J.~Claessens} is with REstore, 2600 Antwerp, Belgium (bert.claessens@restore.eu).}
\thanks{P. Vrancx is with the AI-lab, Vrije Universiteit Brussel, 1050 Brussels, Belgium (pvrancx@vub.ac.be)}
\thanks{\text{F. Spiessens} is with Vito/EnergyVille, 2600 Mol, Belgium.}
\vspace{-0.65cm}
}

\markboth{}%
{Shell \MakeLowercase{\textit{et al.}}: Bare Demo of IEEEtran.cls for IEEE Journals}

\maketitle

\begin{abstract}
This paper considers a demand response agent that must find a near-optimal sequence of decisions based on sparse observations  of its environment.
Extracting a relevant set of features from these observations is a challenging task and may require substantial domain knowledge.
One way to tackle this problem is to store sequences of past observations and actions in the state vector, making it high dimensional, and apply techniques from deep learning.
This paper investigates the capabilities of different deep learning techniques, such as convolutional neural networks and recurrent neural networks, to extract relevant features for finding near-optimal policies for a residential heating system and electric water heater that are hindered by sparse observations.
Our simulation results indicate that in this specific scenario, feeding sequences of time-series to an LSTM network, which is a specific type of recurrent neural network, achieved a higher performance  than stacking these time-series in the input of a convolutional neural network or deep neural network. 
\end{abstract}

\begin{IEEEkeywords}
Convolutional networks, deep reinforcement learning, long short-term memory,  residential demand response 
\end{IEEEkeywords}

%
\IEEEpeerreviewmaketitle

\section{Introduction}
\IEEEPARstart{O}{ptimal} control of Thermostatically Controlled Loads (TCLs), such as heat pumps and water heaters, is expected to play a key role in the
application of residential demand response~\cite{mathieu2015arbitraging,Dupont}. 
TCLs can use their thermal inertia, e.g. a water buffer or building envelope, as a thermal battery to store energy and shift energy consumption in response to changes in the electricity price or to provide grid services.
Amongst the more important challenges hindering the application of residential demand response is partial observability of the environment~\cite{vetteos12,kazmi,OldewurtelChallenges}, where a part of the state remains hidden from the agent due to sensor limitations, resulting in a partially observed control problem.

Model-predictive control (MPC)~\cite{camacho2004model} and Reinforcement Learning (RL)~\cite{sutton1998reinforcement} are two opposing paradigms to solve the optimal control problem of TCLs. 
As such, MPC and RL have developed a set of different techniques to tackle the problem of planning under partial observability.

In MPC, a Kalman filter is often used to  estimate hidden features by exploiting information about the system dynamics and using Bayesian interference.
For example, in~\cite{vetteos12}  Vrettos \textit{et al.} applied a Kalman filter to estimate the temperature of a building envelope and in~\cite{kazmi} Kazmi \textit{et al.} applied a similar approach  to estimate the state of charge of an electric water heater.

RL approaches, on the other hand, store sequences of past interactions with their environment in a memory and extract relevant features based on this memory.
The challenges herein is to consider a priori how many interactions are important to learn a specific task and what exact features should be extracted.
Deep neural networks or multi-layer perceptrons are the quintessential technique for automatic feature extraction in RL~\cite{bertsekas1996neuro,DeepLearningBook}.
An important breakthrough of automatic feature extraction using deep learning is presented in~\cite{deepmind}, where  Mnih \textit{et al} apply a convolutional neural network to automatically extract relevant features based on visual input data to successfully play Atari games.

Finally, by combining RL and MPC, the authors of~\cite{PLATO} presented a method that trains complex
control policies with supervised learning, using MPC to generate the supervision.
The teacher (MPC) uses a rough representation of its environment and full state, and the learner updates its policy based on the partial state using supervised learning.


\section{Literature review}
This section provides a short literature overview of Reinforcement Learning (RL) related to demand response and discusses some relevant applications of deep learning in RL.

\subsection{Reinforcement learning and demand response}
An important challenge in tackling residential Demand Response (DR) is that any prior knowledge in the form of a physical model of the environment and  disturbances is not readily available or may be too costly to obtain compared to the financial gains obtained with DR.
As RL techniques can be applied ``blind'' and consider their environment as a black box, they require no prior knowledge nor do they require a system identification step, making them extremely suited for residential DR.
As a result,  residential DR has become a promising application domain for RL~\cite{WenZhen,kara2012using,Mocanu2016646,RuelensBRLDevice,DDR,franccois2016deep,Mevludin}.
The most important RL algorithms applied to DR are temporal difference RL, batch RL and more recently deep RL.
The first application of RL to demand response were standard temporal difference methods, such as Q-learning and SARSA~\cite{sutton1998reinforcement}.
For example, in~\cite{WenZhen}, Wen~\textit{et al.} showed how Q-learning can be applied to a residential demand response setting and
in~\cite{kara2012using}, Kara~\textit{et al.} applied Q-learning to provide short-term ancillary services to the power grid by using a cluster TCLs.
Extending  this work, Mocanu~\textit{et al.} demonstrated how a deep belief network can be integrated in Q-learning and SARSA to extract relevant features~\cite{Mocanu2016646}, allowing for cross-building transfer learning.

In~\cite{RuelensBRLDevice}, the authors demonstrated how batch RL can be tailored to a residential demand response setting using a set of \textit{hand-crafted} features, based on domain specific insights.
The authors extended a well-known batch RL algorithm, fitted Q-iteration, to include a forecast of the exogenous variables and demonstrated that it outperformed standard temporal difference methods, resulting in a learning phase of approximately 20-30 days, suggesting that batch RL techniques are more suitable for demand response. 

More recently, inspired by advances in deep learning, the authors extended this approach for a cluster of TCLs using an automatic feature extraction method based on convolutional neural networks~\cite{DDR}.
A binning algorithm is used to map the full state of the cluster to a two-dimension representation that can be used as input for the convolution neural network.
Similarly, in~\cite{franccois2016deep}, Fran{\c{c}}ois-Lavet \textit{et al.} applied a convolutional neural network as a function approximator within RL to capture the stochastic behavior of the load and renewable energy production in a microgrid setting with a short-term and long-term storage.

\subsection{Recurrent neural networks and partial observability}
In contrast to vanilla neural networks,  Recurrent Neural Networks (RNNs) have an internal state, which is based on the current input state and the previous internal state, allowing the internal state to act as a memory modeling the impact of previous input states on the current task.
This internal state allows the RNN to process sequences of input data, making it a natural framework to mitigate the problem of partial state information.

In practice, however, RNNs have difficulties learning long-term dependencies~\cite{bengio1993problem}.
An LSTM network is a special type RNN developed by Hochreiter and Schmidhuber in~\cite{LSTMs} that solves the long-term dependency problem, by adding special structures called \textit{gates} that regulate the flow of information to the memory state.

The application of a RNN within Q-learning, called recurrent-Q, was introduced by Lin and Mitchell in \cite{lin1993reinforcement}, demonstrating  that recurrent-Q was able to learn non-Markovian tasks. 
Extending on this idea, Bram Bakker~\cite{bakker2002reinforcement} demonstrated  how LSTM  using advantage learning  can solve non-Markovian tasks with long-term temporal dependencies.
In addition to value-based RL, a successful implementation of a policy gradient method with an LSTM architecture to a non-Markovian task  can be found in~\cite{wierstra2007solving}.
Motivated by the promising results of Deepmind with Deep QN~\cite{deepmind}, the authors of~\cite{flickerAtari} demonstrated how an LSTM network can be combined with a deep Q-network for handling partial observability in Atari games, induced by flickering game screens.


\subsection{Contributions}
This paper investigates the effectiveness of different deep learning techniques within reinforcement learning for demand response applications that are hindered by sparse observations, making the following contributions. 
We present how  an LSTM network, Convolutional Neural Network (CNN) and multi-layer neural network, can be used within a well-known batch RL algorithm, fitted Q-iteration, to approximate the Q-function, extending the state with historic partial observations.
We demonstrate their performance for two popular embodiments of flexible loads, namely a heat pump for space heating and an electric water heater.
The paper is structured as follows. Section \ref{MDP} states the problem and formalizes it as a  Markov decision process. 
Section \ref{LSTM} explains how these deep learning techniques can be used to extract relevant features based on sequences of observations and used within a batch RL. Section~\ref{DL} describes the different deep learning architectures.
Section\ref{SIM} presents the simulation results of two flexibility carriers and finally Section\ref{CON} draws conclusions and discusses further work.

\section{Markov decision-making formalism}
\label{MDP}
This section states the problem and presents the formalism to tackle it.

\subsection{Problem statement}
In most complex real-world problems, such as demand response, an agent cannot measure the exact full state of its environment, but only a partial observation of the state.
Depending on how good this partial observation can be used to model future interactions, using partial state information may result in sub-optimal policies.
This paper presents two demand response applications that are hindered by partial observability, where the agent cannot measure the state directly, but has to extract relevant features based on how much energy the application consumed and how much it lost.
In our  first experiment, we consider a heat-pump agent that can only measure its electricity consumption and outside temperature. 
In the second experiment, we consider an electric water heater agent with partial state state information, consisting of its measured electricity consumption and the flow rate and temperature of the tap water exiting the water buffer.

To tackle this challenge, we will first formalize the underlying problem as a Markov decision process and then introduce the concepts of partial state information.

\subsection{Formalism}

At each discrete time step $k$, the \textit{full} state of the environment evolves as follows: $\bm{x}_{k+1} = f(\bm{x}_{k},u_{k},\bm{w}_{k})$ with $\bm{w}_{k}$  a realization of a random disturbance drawn from a conditional probability  distribution $\probwk$ and
$u_{k} \in U$ the control action.
Associated with each action of the agent, a cost $c_{k}$ is provided by $
c_{k}=\rho(\bm{x}_{k},u_{k},\bm{w}_{k})$, where $\rho$ is a cost function that is a priori given.

The goal of the agent is to find an optimal  control policy ${h^{*}:X\rightarrow U}$ that minimizes the expected $T$-stage return for any state in the state space.
Value-based RL techniques characterize the  policy $h$ is  by using  a state-action value function or Q-function:
\begin{equation}
Q^{h}(\bm{x},u) = \underset{\bm{w}\sim\probwk}{\expected} \left[\rho(\bm{x},u,\bm{w}) + J^{h}_{T}(f(\bm{x},u,\bm{w})) \right],
\label{Qfunction}
\end{equation}
The Q-function is the cumulative return starting from  state $x$, taking action $u$, and following $h$ thereafter.
Given the Q-function, an action for each state can be found as:
\begin{equation}
h(\bm{x}) = \text{arg min~} Q^{h}(\bm{x},u).
\label{Qfunction}
\end{equation}
This paper applies a value-based batch RL technique to approximate the Q-function corresponding to the optimal policy  based on an imperfect observation of the true state.

\subsection{Partial state}
It is assumed that the state space $X$  measured by the agent consists of three components: timing-related state information $X^{\timing}$, controllable state information $X^{\physical}$, and exogenous (uncontrollable) state information $X^{\exo}$. 
In this work the timing related is given by the current quarter in the day $\xtimingk \in X^{\timing} = \left\{1,\dots,96\right\}$, which allows the agent to capture time-varying dynamics.
The controllable state information $\xphys \in X^{\physical}$ comprises the operational 
measurements that are influenced by the control action.
In reality, mosts agent can only measure a partial observation $\obsk$ of the true state $\xphys_k$, resulting in a partially observable Markov decision problem.
The exogenous information $\xexok$ is invariant for control actions ${u_{k}}$, but has an impact on the dynamics. Examples of exogenous variables are weather conditions and demand profiles (e.g heat demand).

Thus, the state measured by the agent at step $k$ is given by:
\begin{equation}
\obsx_k = (\xtimingk,\obsk,\xexok).
\label{partial_state}
\end{equation}
Note that since (\ref{partial_state})  only includes part of the true state, it becomes impossible to model future state transitions, making the state non-Markovian.

\subsection{Action}
\label{subsection.backup_controller}
At each time step, a  demand response agent can request an action $u_t\in [0,1]$: either to switch OFF or ON.
To guarantee the comfort and safety constraints of the end users, each TCL is equipped with an overrule mechanism (or thermostat).
The backup function  ${B:X \times U \longrightarrow \Uphys}$ maps the requested control action $u_{k} \in U$ taken in  state $\bm{x}_{k}$ to a physical control action $\uphk \in \Uphys$: 
\begin{equation}
\uphk= B(\bm{x}_{k},u_{k}).
\end{equation} 
The settings of the backup function $B$ are unknown to the learning agent, but the resulting action  $\uphk$  can be measured by the learning agent.

\subsection{Cost}
This papers considers a dynamic pricing scenario where an external price profile is known deterministically at the start of the  optimization horizon:
\begin{equation}
c_{k}= \rho(\uphk,\lambda_k) = \uphk \lambda_k \Delta t,
\label{eq_ToU}
\end{equation}
where $ \lambda_k$ is the electricity price at time step $k$ and $\Delta t$ is the length of a control period.

\section{Batch reinforcement learning}
\label{LSTM}
Given full observability, batch RL algorithms start with a batch of four tuples of the form:  $\left(\x_{k},u_{k},\x_{k}', \uphysk \right)$,  where $\x_{k}$ represents the true state of the problem.

According to the theory of partial observable Markov decision processes~\cite{bertsekas1996neuro}, the optimal value function at time step $k$ depends on the partial state observations of \textit{all} proceeding periods.
However, since these observations accumulate over time, it is important to capture sufficient statistics, thats  a history length $h$  which summarizes the essential content of the measurements.
As such, this paper tackles the problem of partial observability by augmenting the state vector with a sequence of partial state observation, requested actions and physical actions of the last $h$ observations:
\begin{equation}
\augx_k = \big(\xtimingk, \histx_{k},\xexok \big)
\label{aug1}
\end{equation}
with $\histx_{k}$ give by:
\begin{equation}
[\obsk,\ldots, \obs_{k-h}],[\uphys_{k-1}, \ldots, \uphys_{k-1-h}],[u_{k-1},\ldots, u_{k-1-h}].
\label{aug2}
\end{equation}
As a results, this paper starts from a bath of four tuples given by:  $ \{(\augx_{k}, u_k, \augxNext_{k}, \uphysk)\}_{k=1}^{\mathcal{M}}$, where $\augx_{k}$ represents the augmented state.
An important challenge is to learn how to extract relevant features from  in a scalable way.

\subsection{Fitted Q-iteration}
This paper applies  fitted Q-iteration~\cite{ernst2005tree} to obtain an approximation of the Q-function $\widehat{Q^{*}}(\augx,u)$.
Fitted Q-iteration iteratively approximates the Q-functions for each state-action pair using its corresponding cost and  the approximation of the Q-function from the previous iterations.
To leverage the availability of forecasts of exogenous information , e.g. outside temperatures, we use the extension of fitted Q-iteration as presented in~\cite{RuelensBRLDevice}, which replaces the observed exogenous
information by its forecasted value $\xhatexolNext$  (line \ref{use_forecast_exo} in Algorithm~\ref{fqilstm}).

%

In order for  Algorithm~\ref{fqilstm} to work, we need to select an approximator architecture  (step~\ref{Qfitting}) that is able to learn relevant features from sequences of input data and  that can generalize the Q-function.

\begin{algorithm}[t]
 	\caption{Batch RL~\cite{ernst2005tree} using LSTM~\cite{LSTMs}}
 	\label{fqilstm}
 	\begin{algorithmic}[1] 
 		\algsetup{linenosize=\tiny}
 		\renewcommand{\algorithmicrequire}{\textbf{Input:}}
 		\REQUIRE $\mathcal{F}=\{\obsx_{k}, u_k, \obsxNext_{k}, \uphysl\}_{k=1}^{\#\mathcal{F}}$ , $[\lambda_k]_{l=1}^T$, $[\xhatexol]_{l=1}^T$, history length $h$\\
 		\STATE Construct $\mathcal{M}=\{\augx_{k}, u_k, \augxNext_{k}, \uphysl\}_{k=1}^{\#\mathcal{M}}$ using (\ref{aug1}) and (\ref{aug2})
 		\STATE let $\widehat{Q}_{0}$ be zero everywhere on $X$ $\times$ $U$
 		\FOR {$N=1,\ldots,T$}
 		\FOR {$k = 1,\ldots,\#\mathcal{M}$}
 		\STATE $~~l \leftarrow \xtiming_{k}$
 		\STATE $~~c_{k} \leftarrow \uphysk \lambda_{l} \Delta t$
 		\STATE $~~\augxNext_{k} \leftarrow (\xtimingnext_{k},\histxnext_k,\xhatexolNext)$
        \label{use_forecast_exo}
 		\STATE $~~Q_{N,k}\leftarrow c_{k} +\underset{u \in U}{\text{min~}}\widehat{Q}_{N-1}(\augxNext_{k},u)  $  
 		\ENDFOR
 		\STATE use approximator (Fig.~\ref{Qapproximator}) to obtain $\widehat{Q}_{N}$ from $\mathcal{T} = \left\{\left((\augx_{k},u_{l}),Q_{N,k}\right),k =1,\ldots,\#\mathcal{M}\right\}$ \label{Qfitting}
 		\ENDFOR
 		\ENSURE $\widehat{Q}^{*}=\widehat{Q}_{N}$
 	\end{algorithmic}
\end{algorithm}

\begin{figure*}[t]
	\centerline{\includegraphics[width=16cm]{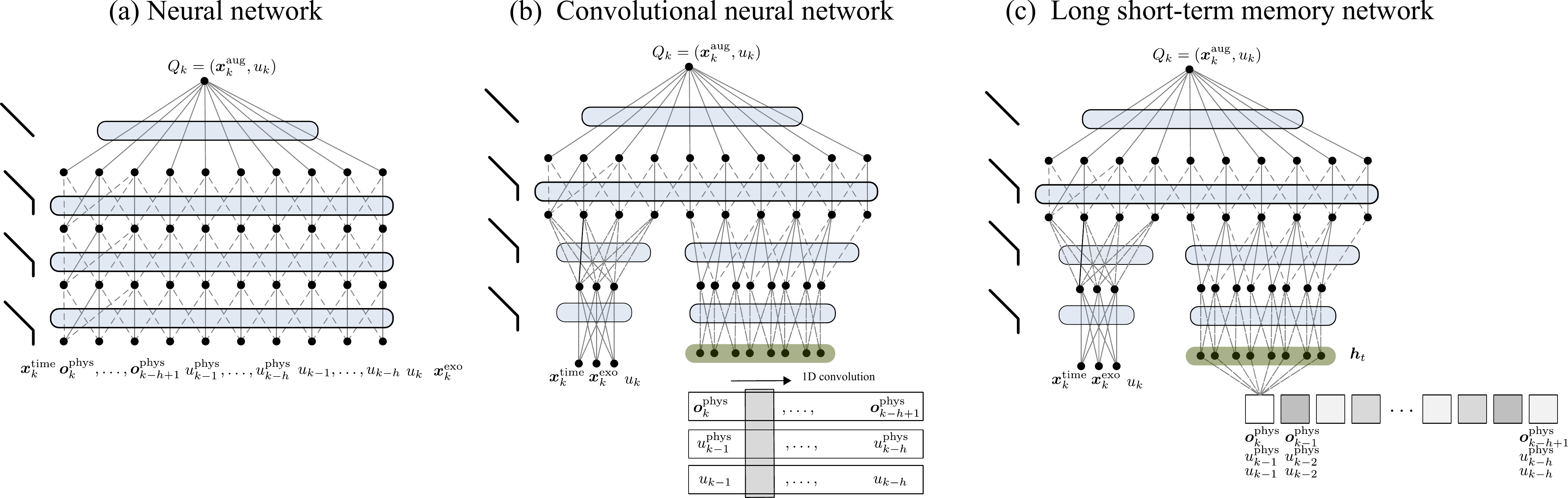}}
	\caption{Sketch of the deep learning architectures used in the simulation section. The LSTM network is represented as an unfolded computational graph, where each node is associated with one particular time instance.}
	\label{Qapproximator}
\end{figure*}

\section{Deep learning approximators}
\label{DL}
This paper investigates the effectiveness of the following deep learning approximators when combined with fitted Q-iteration.

\subsection{Deep neural network }
It has been shown that a neural network with a single layer is sufficient to represent any function, but the layer may become infeasible large and  may fail  to train and generalize correctly.
To overcome these two challenges, deeper networks are used as these networks can reduce the number of units to represent the function and can reduce the generalization error.   
Fig. ~\ref{Qapproximator}(a) illustrates the neural network as used in this paper, consisting of an input layer given by ($\augx_k,u_k$), two hidden layers with rectified linear units (ReLUs), and one linear output layer, representing the approximated Q-function. 

\subsection{Convolutional neural network}
CNNs have been successfully applied to extract features from image data, represented as a 2D grid of pixels.
In this paper, we consider  a time series and convolve a 1D filter of length N over the time-series in the state (\ref{aug2}).
A sketch of the applied CNN can be seen in Fig.~\ref{Qapproximator}(b), which consists of two components that are merged to output a singe value. The first component is a dense neural network which takes the timing-related information,  exogenous information and action as input.
The second components is a CNN which takes the time-series as input (\ref{aug2}).
For each sequence, the network consists of two layers containing eight 1D filters of length $4$ followed by a ReLU, which is downsampled by using an average pooling layer.

\subsection{Long short-term memory}
\subsubsection{Background}
An LSTM network (Fig.~\ref{Qapproximator}) consists of LSTM nodes that are recurrently connected to each other.
Each LSTM node has an internal recurrence or memory cell $C^{(t)}$ and a system of gating units that controls the flow of informations.
 For each step $t$ of the sequence $\xlstm^{(1)},\ldots,\xlstm^{(t)},\ldots,\xlstm^{(h)}$, the  resulting action of the forget gate $\bm{f}^{(t)}$, input gate $\bm{i}^{(t)}$ and output gate $\bm{o}^{(t)}$ of a single LSTM nodes is provided by:
 \begin{equation}
 \begin{split}
 &\bm{f}^{(t)} = \sigma(W_{f}[\bm{h}^{(t-1)},\xlstm^{(t)}] + \bm{b}_f),\\
 &\bm{i}^{(t)} = \sigma(W_{i}[\bm{h}^{(t-1)},\xlstm^{(t)}] + \bm{b}_i),\\
 &\bm{o}^{(t)} = \sigma(W_{o}[\bm{h}^{(t-1)},\xlstm^{(t)}] + \bm{b}_o),
 \end{split}
 \end{equation}
 where $W_{f},W_{i},W_{o}$ and $\bm{b}_{f},\bm{b}_{i},\bm{b}_{o}$ are the weights and biases of the forget, input and output gate, $\sigma$ denotes the logistic sigmoid function and $\xlstm^{(t)}$ denotes the current element of sequence~(\ref{sequence}), with the time step index $t$ ranging from 1 to $h$.

 The  internal memory cell of the LSTM node is updated as:
 \begin{equation}
 \bm{C}^{(t)} = \bm{f}^{(t)}*\bm{C}^{(t-1)} + \bm{i}^{(t)} *\widetilde{\bm{C}^{(t)}},
 \end{equation}
 where $\widetilde{\bm{C}^{(t)}}$ and $\bm{C}^{(t-1)}$ are the current and previous memory  state and $*$ denotes a pointwise multiplication operator.
 Note that the new memory $\bm{C}^{(t)}$ is defined by the information it forgets from the old state $\bm{f}^{(t)}* \bm{C}^{(t-1)}$ and remembers from the current $\bm{i}^{(t)}* \widetilde{\bm{C}^{(t)}}$.
    
In a last step, a hyperbolic tangent function is applied to the memory cell and multiplied with the output $\bm{o}^{(t)}$, which defines what information to output.
  \begin{equation}
 \bm{h}^{(t)} = \bm{o}^{(t)}* \text{tanh}(\bm{C}^{(t)}),
 \end{equation}

This gating mechanism allows the LSTM network to store information about the state for long periods of time and protects the gradient in the cell from harmful changes during training related to the vanishing or exploding gradient problem of RNN~\cite{bengio1993problem}.


\subsubsection{Approximator architecture}
The approximator architecture consists of two components: an LSTM network and a standard multi-layer perceptron (Fig.~\ref{Qapproximator}).
The first part of the input, corresponding to the LSTM component, contains the historic information of the partial state $\histx$.
For each $k=h+1,\ldots,\#\mathcal{M}$, the input of the LSTM network is given by the following sequence:
\begin{equation}
\underbrace{\begin{bmatrix}
\obs_{k}\\ 
\uphys_{k-1}\\ 
u_{k-1}
\end{bmatrix}}_{\xlstm^{(1)}_k}
,\ldots,
\underbrace{\begin{bmatrix}
\obs_{k-h+1}\\ 
\uphys_{k-h}\\ 
u_{k-h}
\end{bmatrix}}_{\xlstm^{(h)}_k}
\label{sequence}
\end{equation}
The history depth $h$ defines how much time steps the network can see in the past to compute its approximation of the Q-function.
The length of the memory cell $d_{\text{cell}}$ represents an important hyper parameter and defines how many knowledge can be encoded. 
 As can be see in Fig.~\ref{Qapproximator} only the content of the last memory cell $h_t$ is used as an input for the next layer.
 
The second part contains the time-related information, exogenous information and action: $\xtimingk, \xexok, u$ .
The outputs of both components are combined to form a singe architecture, which is followed by two fully connect layers with Rectified Linear Unit (ReLU) activation functions.
A final linear output layer approximates the final Q-function for the provided state-action pair.  


\section{Simulation experiments}
\label{SIM}
This section evaluates the performance of combining the presented deep learning techniques with Alogirthm 1 for two  providers of demand flexibility  exposed to a dynamic energy price.

\subsection{Simulation framework}
At the end of each simulation day, Algorithm~\ref{fqilstm} is used to compute  a new policy based on current batch and electricity price for the following day. 
The RL  agent starts with an empty batch and alternates exploration and exploitation according to a decreasing exploration probability: $\varepsilon_d = 0.75^d$, where $d$ denotes the current episode.

All experiments are repeat 10 times starting form a different random seed, resulting in different exploration probabilities and stochastic disturbances. The following results indicate the average of these simulation experiments, where a confidence bound ($\pm2\sigma$) is indicated by a shaded area, representing a 0.95  probability that the solution lies in the shaded area.

The average simulation time for one day (Algorithm~\ref{fqilstm}) is about 1.5 hour\footnote{Simulation hardware: Xeon E5-2680 v2 processor with 15 GiB memory (Amazon elastic cloud instance type: c3.2xlarge).} using Keras with Theano as backend.

\subsection{Experiment 1: Space heating}
Similar as in~\cite{zhang2012aggregate,DDR}, a second-order heat-pump model ($C_a=2.441\mathrm{MJ}/^\circ\mathrm{C}, U_a=125\mathrm{W}/^\circ\mathrm{C}, C_m = 9\mathrm{MJ}/^\circ \mathrm{C}, H_m=6.863\mathrm{kW}/^\circ\mathrm{C}$) with outside temperatures from~\cite{ukkel} is used to simulate the temperature dynamics of a residential building with a heat pump. The heat pump has a maximum electric heating power of 2.3kW and the minimum and maximum comfort settings are set to $20^{\circ}C$ and $23^{\circ}C$.
To model stochastic impact of user-behavior we  sample an exogenous temperature disturbance from $\mathcal{N}(0,0.025)$.
The time resolution of the dynamics is 60 seconds and of the control policy is 15 minutes. 

The state vector describing the environment is defined as:
\begin{equation}
\x_k = \big(\xtimingk,\Ta, \Tm, \Tout, \Texo \big),
\label{fullstate}
\end{equation}
where $\xtiming_k$ contains timing information, $\Ta$ the air temperature, $\Tm$ the virtual mass temperature, $\Tout$ the outside temperature and $\Texo$ an exogenous disturbance.
As stated in the problem description, it is assumed that the RL agent cannot measure the air and mass temperature of the building, resulting in a partial observed control problem.
As such, we construct the following augmented state vector:
\begin{equation}
\begin{split}
\augx_k = \big(\xtiming_k, &[\uphys_{k-1},\ldots, \uphys_{k-h}],\\
                   &[u_{k-1},\ldots, u_{k-h}], \\
                   &[\Tout_{k-1},\ldots, \Tout_{k-h}], \Tout_k \big),
\end{split}
\label{sequenceNN}
\end{equation}
which includes three time series of lenght $h=20$. 


\subsubsection{NN Architecture}
The neural network consists of three dense layers with 50 neurons with ReLU activation functions, followed be a linear output unit. The neural network was trained using RMSprop with a minibatch size of 32.

\subsubsection{CNN Architecture}
The network consists of two components, namely a CNN and dense network.
The CNN  component consists of two 1D convolutions (along the time dimension) that are each followed by an average pooling layer.
The dimension of the first filter is $(L\times3)$, where $L$  is the filter length and 3 is number of input sequences and the dimension of the second filter is $L\times1$. Both filters have a filter length of 4.
The dense network processes the time-related information, exogenous information and action.
Both components are merged and followed with two layers with 20 neurons and a single output layer. 
All layers use ReLU activation function except for the output layer that uses a linear function.

\subsubsection{LSTM Architecture}
The input to the LSTM network is provided by the sequence: 
\begin{equation}
\begin{bmatrix}
\uphys_{k-1},\\
u_{k-1},\\
\Tout_{k-1}
\end{bmatrix}
 ,\ldots, 
  \begin{bmatrix}
 \uphys_{k-h},\\
 u_{k-h},\\
 \Tout_{k-h}
\end{bmatrix}
\end{equation}
 and the NN is provided by $\xtiming_k, \Tout_k$ and $u_k$.
For the heatpump experiment the best results were obtained with the history depth $h$ set to 20 time steps (quarters) and the length of each LSTM memory cell $d_{\mathrm{cell}}$ set to 8. 

\subsubsection{Convergence}
\begin{figure}[t!]
	\centerline{\includegraphics[width=8.0cm]{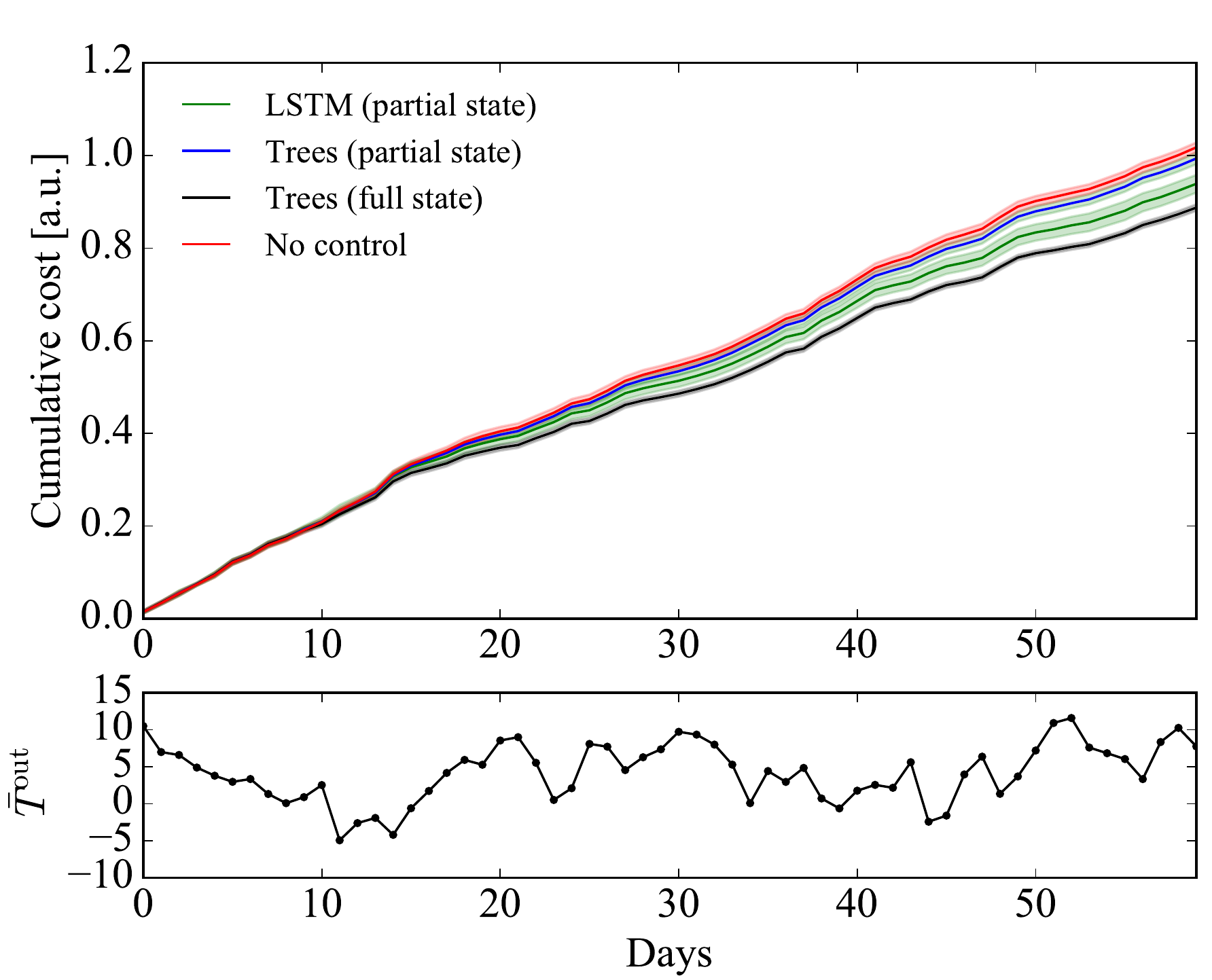}}
	\caption{Top: cumulative cost of the heat pump experiment using FQI-LSTM and FQI-Trees. Bottom: corresponding daily average outside temperature.}
	\label{hp_convergence}
\end{figure}

Fig.~\ref{hp_convergence} depicts the cumulative cost using function approximators (top) and daily average outside temperature (bottom). The \textit{no control} strategy activates the backup controller, without setting a control action, and can be seen as a worst case scenario as it is agnostic about the electricity price.
An upper bound is computed by considering the full state information as defined in~(\ref{fullstate}).
In addition to LSTM with partial state information, the figure depicts the  cumulative of using an ensemble of extremely randomized trees (or ExtraTrees)~\cite{ernst2005tree}.
The number of trees in the ensemble was set to 100 and the minimum sample size for splitting a node to 5.
Our results indicate that the ExtraTrees approximator was not able to extract relevant features from the partial state information and performed only $1.5\%$ better than the  no control strategy.
In contrast, the LSTM  approximator was able to extract relevant features and achieved a reduction of $5.5\%$.
 
\begin{figure}[t!]
	\centerline{\includegraphics[width=9cm]{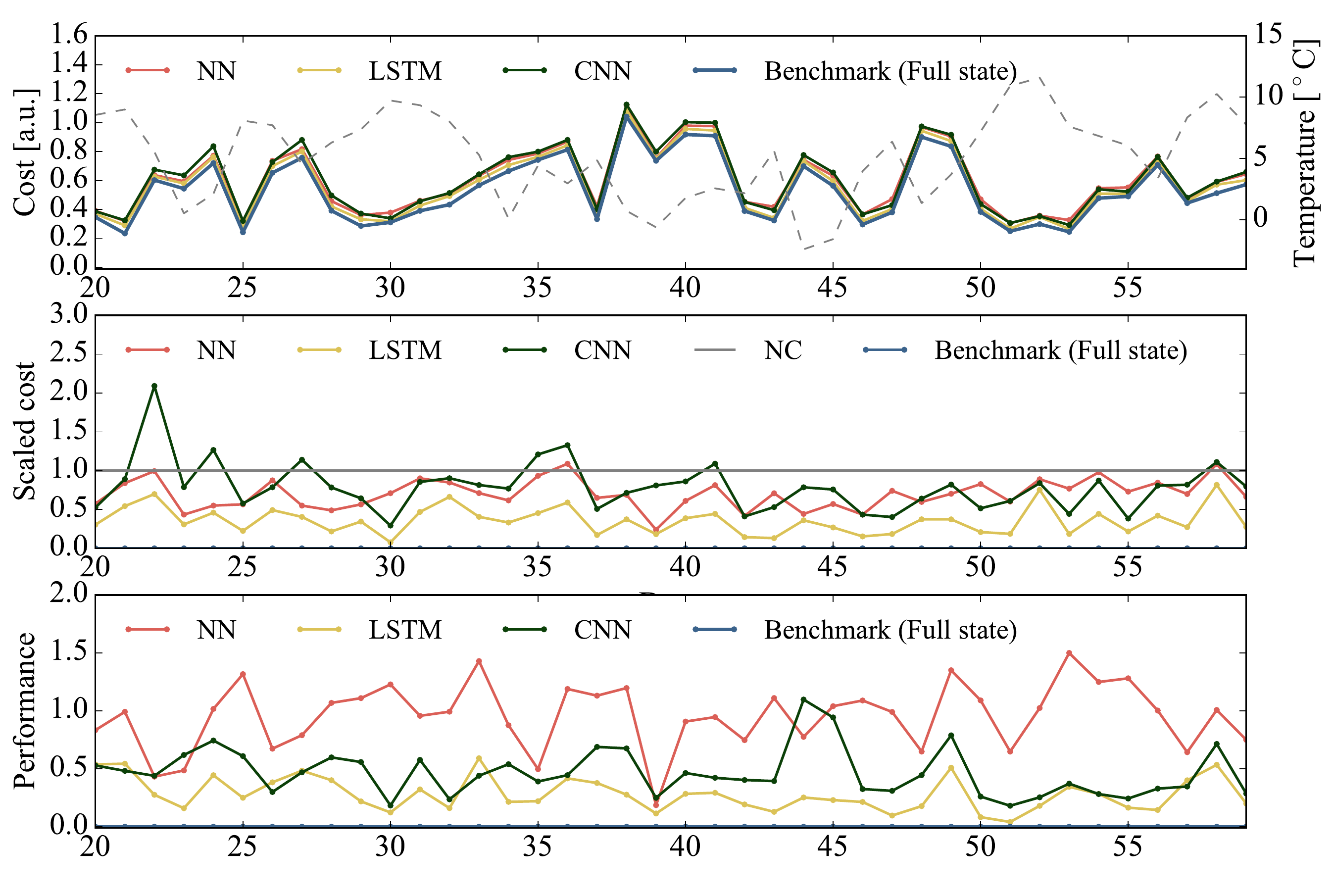}}
	\caption{Top: daily cost for the heat pump experiment using FQI-NN, FQI-LSTM and FQI-CNN based on sparse observations. Middle: corresponding scaled daily cost. Bottom: metric defined by the distance between the near-optimal policy (benchmark) and policy obtained with FQI-NN, FQI-LSTM and FQI-CNN.}
	\label{hp_scaled_convergence}
\end{figure}

Fig.~\ref{hp_scaled_convergence} shows the daily cost (top) obtained with Algorithm~\ref{fqilstm}, using a partial state information using a neural network, CNN and LSTM network as a function approximator.
The middle graph indicates the scaled cost which is $c$ calculated as follows: $(c - c_{\mathrm{f}})/(c_{\mathrm{nc}} - c_{\mathrm{f}})$, where $c_{\mathrm{f}}$ is the result of using the full state information and $c_{\mathrm{nc}}$ of using the no control strategy, resulting in $c=0$ for the full state strategy and $c=1$ for the no control strategy.
This figure indicates Algorithm 1 obtained a scaled cost of  $0.37$ using LSTM, $0.66$ using NN and $0.82$ using CNN.
The bottom graph compares the resulting control policies of LSTM, CNN en NN with the control policy of the full state using a euclidean distance. 
Although NN achieved a better performance than CNN, the resulting policy of CNN and LSTM are closer to the policy of the full state.
We speculate that the CNN and LSTM learned a better representation of the full state than the NN, since the NN achieved a low cost by lowering the air temperature to minimum temperature without reacting to the price.

\subsubsection{Daily results}
\begin{figure*}[t!]
	\centerline{\includegraphics[width=15cm]{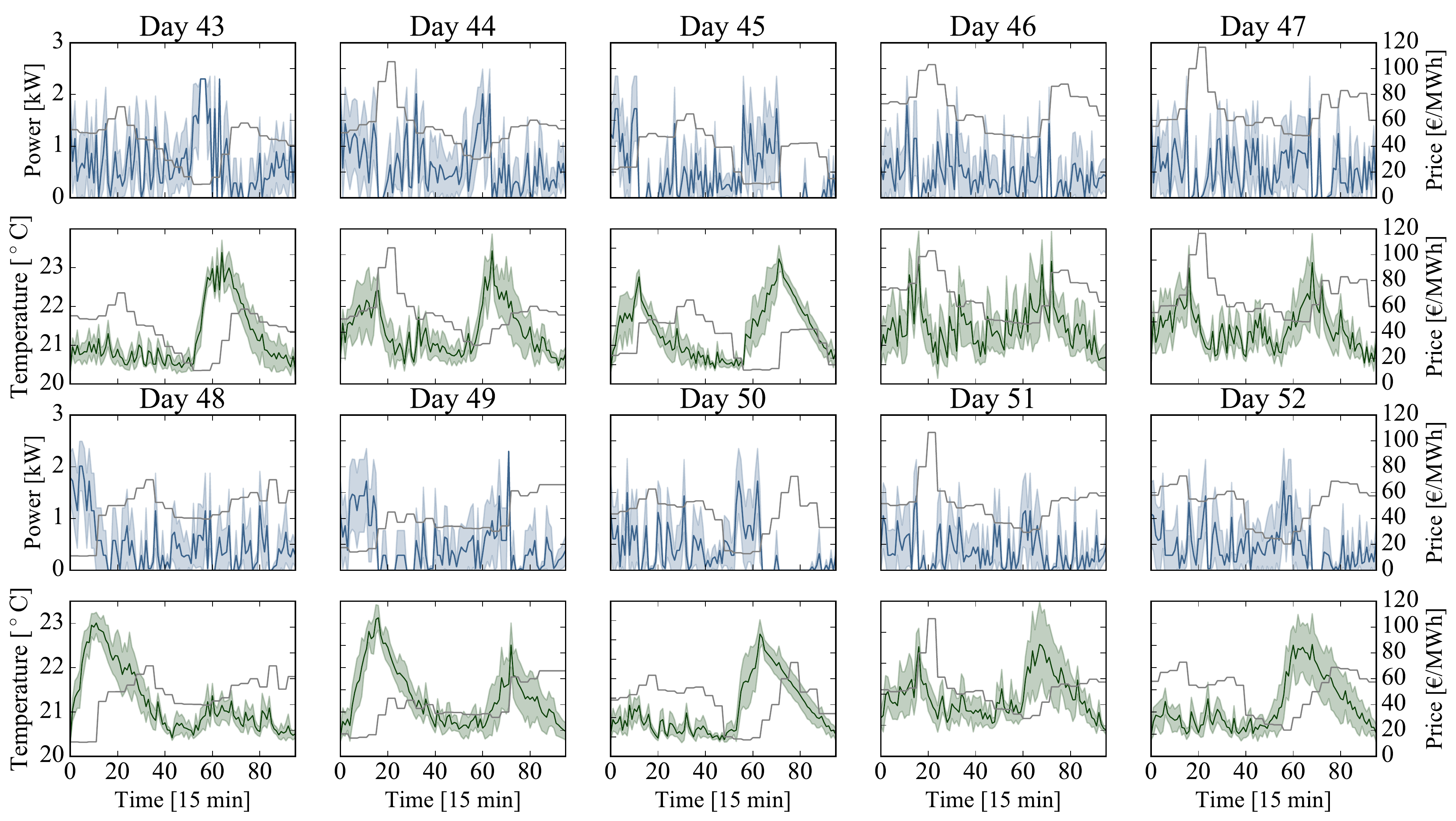}}
	\caption{Power consumption (first and third row) and air temperatures (second and fourth row) for 10 greedy simulation days (left y-axis) using FQI-LSTM with partial state information  for the heat pump experiment. The corresponding price profiles are depicted in gray (right y-axis).}
	\label{heatpump_sim}
\end{figure*}

A more qualitative interpretation of our results can be seen in Fig.~\ref{heatpump_sim}.
The figure shows the power consumption and the corresponding  daily price profiles. It can be seen that the learning agent successfully postponed its power consumption to low price moments, while satisfying the comfort constraints. 

\subsection{Experiment 2: Electric water heating}
The second experiment considers an electric water heater with a water buffer of 200 liters and a daily average water consumption of 100 liter. The minimum and maximum water temperature is set to $45^{\circ}C$ and $65^{\circ}C$. The water heater is equipped with a thermostatic mixing value to assure a constant requested temperature of $45^{\circ}C$. The water heater has an electric power rating of 2.3kW and a built-in backup controller as defined in~\cite{SMARTBOILER}.
The time resolution for the dynamics is  5 seconds and the time resolution for the control policy is 15 minutes. 

The full state vector of the electric water heather is defined by:
\begin{equation}
\x_k = \big(\xtiming_k,T_{k}^{1},\ldots,T_{k}^{|\mathcal{L}|}, d_k),
\end{equation}
where $T_{\sens,t}^{i}$ is the temperature corresponding to the $i$th layer and $d_k$ is the current tap demand.
During our simulation, a non-linear stratified model with 50 layers is used to simulate the temperature gradient along the water tank and stochastic tap water profiles are used based on~\cite{SMARTBOILER}.

In a previous paper~\cite{kak_boiler}, the authors considered that the agent could measure a imperfect state through eight temperature sensors.
In this experiment, however, it is assumed that the buffer is not equipped with a set of sensors to measure the different temperatures inside the water buffer. 

As a result, we define the following augmented state vector:
\begin{equation}
\begin{split}
\augx_k = \big(\xtiming_k,&[u_{k-1},\ldots u_{k-h}],[\uphys_{k-1},\ldots \uphys_{k-h}]\\
				  &[\flow_k,\ldots \flow_{k-h+1}] , [T_k^{\topLayer},\ldots, T_{k-h+1}^{\topLayer}] \big),
\end{split}
\end{equation}
where $\xtiming_k$ contains timing information, $u_k$ is the requested control action, $\uphys$ the actual action, and $\dot{m}_k$ and $T_k^{\topLayer}$ are the mass flow rate and temperature of the water exiting the water buffer.
Note that $[\uphys_{k-1},\ldots \uphys_{k-h}]$ represents the electricity consumption of the boiler and $[\flow_k,\ldots \flow_{k-h+1}] , [T_k^{\topLayer},\ldots, T_{k-h+1}^{\topLayer}]$ represents the energy flowing out of the boiler.


\subsubsection{(C)NN Architecture}
The NN and CNN architecture are identical as in the previous experiment with the exception that the filters size of the first convolutional layer is $4\times4$, because now we have 4 input sequences.

\subsubsection{LSTM Architecture}
The input to the LSTM network is provided by the sequence:
\begin{equation}
\begin{bmatrix}
\uphys_{k-1},\\
u_{k-1},\\
\dot{m}_k,\\
T_k^{\topLayer}
\end{bmatrix}
,\ldots, 
\begin{bmatrix}
\uphys_{k-h},\\
u_{k-h}\\
\dot{m}_{k-h+1},\\
T_{k-h+1}^{\topLayer}
\end{bmatrix}
\end{equation}
For the boiler experiment the best results were obtained with 
the history depth $h$ set to 40 time steps (quarters) and the length of each LSTM memory cell $d_{\mathrm{cell}}$ set to 12. 
\begin{figure*}[t!]
	\centerline{\includegraphics[width=15cm]{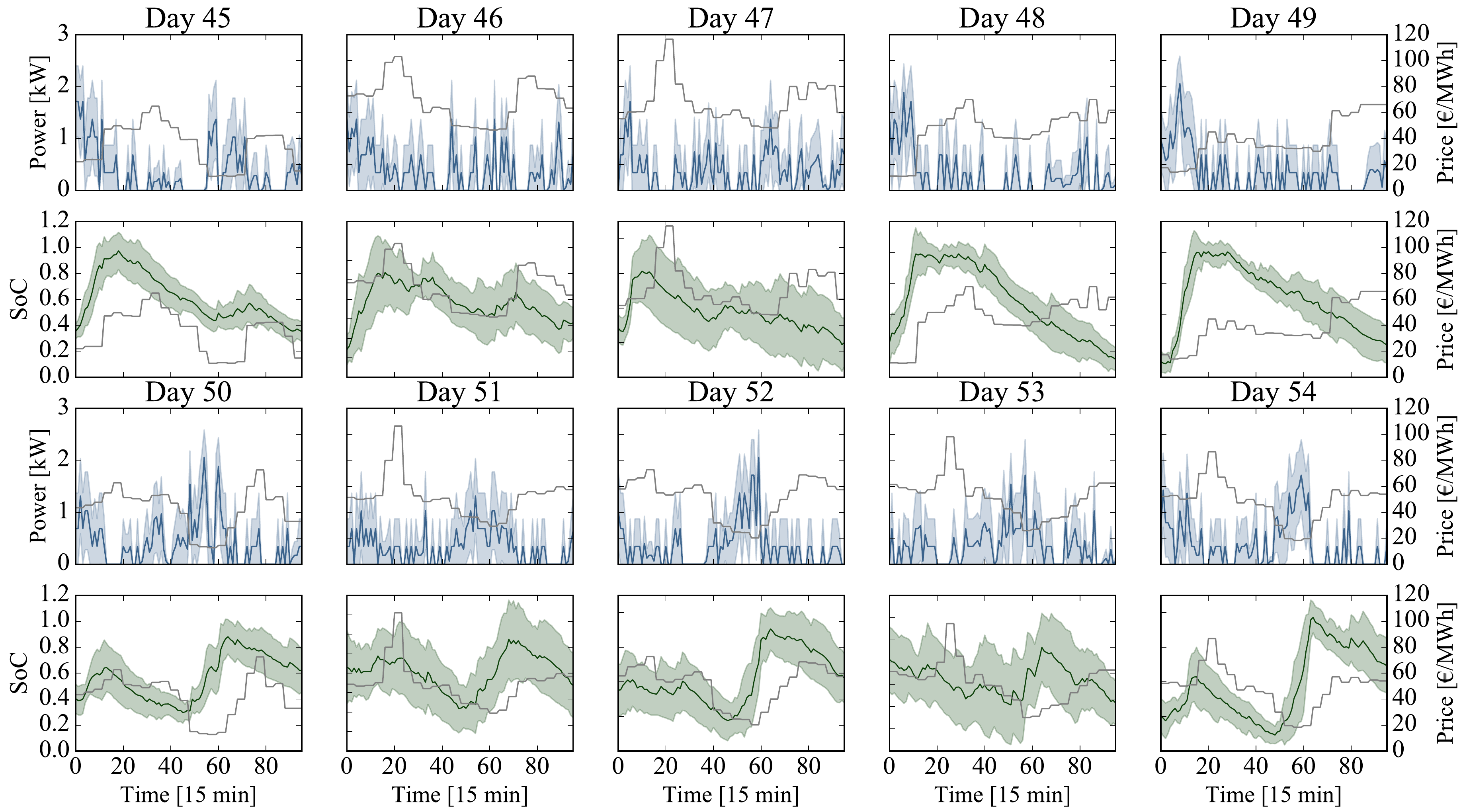}}
	\caption{Power consumption (first and third row) and state of charge (second and fourth row) for 10 greedy simulation days (left y-axis) using FQI-LSTM with partial state information for the electric water heater experiment. The corresponding price profiles are depicted in gray (right y-axis).}
	\label{boiler_sim}
\end{figure*}

\begin{figure}[t!]
	\centerline{\includegraphics[width=9.5cm]{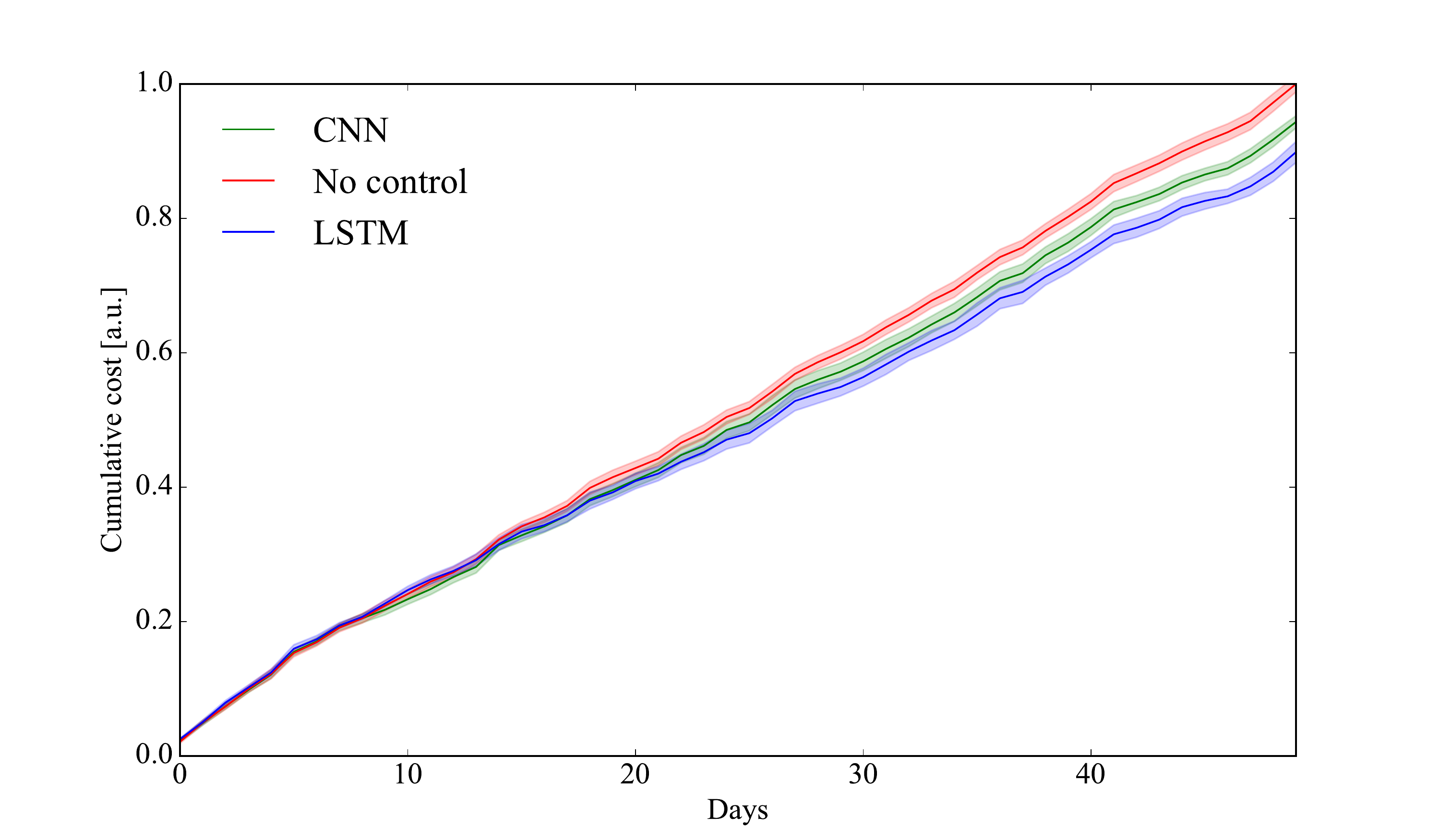}}
	\caption{Cumulative cost of the electric water heater experiment using FQI-LSTM and FQI-CNN based on partial state information.}
	\label{boiler_convergence}
\end{figure}

\subsubsection{Daily results}
For the electric water heater scenario, we only offer qualitative results (Fig.~\ref{boiler_sim}).
It shows the daily power consumption of an  electric water heater and corresponding price profiles. It can be seen that the learning agent required four weeks of learning before obtaining reasonable policies (lower row of graphs).
A final comparison between using a CNN or LSTM network as a function approximator can be seen in Fig.~\ref{boiler_convergence}, indicating that using a CNN resulted in a cost reduction of $5.5\%$ and using a LSTM network in $10.2\%$. The results of FQI-NN were omitted because we were to able to stabilize the learning of the NN.

\section{Conclusions and future work}
\label{CON}
In this paper, we demonstrated the effectiveness of combining different deep learning techniques with reinforcement learning for two demand response applications that are hindered by sparse observations of the true state.
Since these sparse observation result in a non-Markovian control problem, we extended the state with sequences of past observations of the state and action. 
	
In a first experiment, we considered an agent that controls a residential heating system under a dynamic pricing scenario, where the agent can only measure its electricity consumption, control action and outside temperature. 
Our simulations indicated that reinforcement learning with long short-term memory (LSTM) performed better than other techniques such as a  neural network, convolutional neural network and ensemble of regression trees, when sparse observations are used.
In our second experiment, we considered an agent that controls a residential electric water heater with a hot storage vessel of 200 liter.
In this scenario, the agent can only measure its electricity consumption, control action and flow and temperature of the tap water exiting the storage vessel.
The simulation results indicated that the LSTM network outperformed the  convolutional network and deep neural network. 

We speculate that the higher performance of the LSTM network comes from its internal memory cell which can act as an integrator.
This internal memory cell allows the LSTM network to process sequences of sparse observations and extract relevant features from it that can represent the underlying state of charge (or energy level) of the application.
A potential direction of future research would be to extract and visualize the relevant features that are being detected by the LSTM network and CNN that would lead to a better performance and understanding.

\vspace{-0.35cm}

\bibliographystyle{IEEEtran}  
\bibliography{references}

\end{document}